\pdfoutput=1
\documentclass[twocolumn]{article}
\usepackage[T1]{fontenc}

\usepackage{siunitx}
\usepackage{algorithm}
\usepackage{algpseudocode}
\usepackage{hyperref}
\usepackage[capitalise]{cleveref}
\crefname{algorithm}{Alg.}{Algs.}
\usepackage{xcolor}

\usepackage[pdftex]{graphicx}
\graphicspath{{./figs/}}

\usepackage{natbib}

\crefformat{footnote}{#2\footnotemark[#1]#3}

\def\nland{\end{tabular}\\\begin{tabular}[t]{c}}

\usepackage{iidtp}
\iidtpTitle{Low Fidelity Digital Twin\\for Automated Driving Systems:\\Use Cases and Automatic Generation}
\iidtpAuthors{
	\iidtpAuthor{Jiri Vlasak}{https://orcid.org/0000-0002-6618-8152}
	\iidtpAuthor{Jaroslav Klap\'{a}lek}{https://orcid.org/0000-0001-8816-2773}
	\iidtpAuthor{Adam Kollar\v{c}\'{i}k}{https://orcid.org/0000-0002-9150-839X}
	\\
	\iidtpAuthor{Michal Sojka}{https://orcid.org/0000-0002-8738-075X}
	\iidtpAuthor{Zden\v{e}k Hanz\'{a}lek}{https://orcid.org/0000-0002-8135-1296}
}
\iidtpInfo{
	\iidtpInfoURL{Source Code}{https://git.sr.ht/~qeef/gen-sdf}
	\iidtpInfoURL{Video}{https://libre.video/videos/watch/e15b96af-9c2c-4b72-adf8-234148d6044c}
	\iidtpInfoURL{Video}{https://libre.video/videos/watch/b2e90dd6-ef61-4756-822f-a915aa2aa23a}
}
\iidtpBibKey{vlasak_low_2024}
\begin{document}
\makeiidtp
\title{Low~Fidelity Digital~Twin for~Automated~Driving~Systems: Use~Cases and~Automatic~Generation}
\author{
	Jiri Vlasak$^1$$^2$\footnote{\protect\url{https://orcid.org/0000-0002-6618-8152}}
	\and Jaroslav Klap\'{a}lek$^1$$^2$\footnote{\protect\url{https://orcid.org/0000-0001-8816-2773}}
	\and Adam Kollar\v{c}\'{i}k$^1$$^2$\footnote{\protect\url{https://orcid.org/0000-0002-9150-839X}}
	\nland Michal Sojka$^2$\footnote{\protect\url{https://orcid.org/0000-0002-8738-075X}}
	\and Zden\v{e}k Hanz\'{a}lek$^2$\footnote{\protect\url{https://orcid.org/0000-0002-8135-1296}}
	}
\date{{\small
$^1$Faculty of Electrical Engineering, Czech Technical University in Prague
\\
$^2$Czech Institute of Informatics, Robotics and Cybernetics, Czech Technical University in Prague
\\
\{jiri.vlasak.2, jaroslav.klapalek, adam.kollarcik, michal.sojka, zdenek.hanzalek\}@cvut.cz
}}
\maketitle

\begin{abstract}
Automated driving systems are an integral part of the automotive industry.
Tools such as Robot Operating System and simulators support their development.
However, in the end, the developers must test their algorithms on a real vehicle.
To better observe the difference between reality and simulation--the reality gap--digital twin technology offers real-time communication between the real vehicle and its model.
We present a low fidelity digital twin generator and describe situations where automatic generation is preferable to high fidelity simulation.
We validate our approach of generating a virtual environment with a vehicle model by replaying the data recorded from the real vehicle.
\end{abstract}


\section{Introduction}
\label{s:intro}


The development and testing of vehicles equipped with automated driving systems (ADS) is expensive, slow, and complicated.
In many cases, simulation comes to the rescue.
However, setting up the simulator and the virtual entities, i.e., the environment and the models, is usually complicated, slow, and therefore expensive.
The difficulties in setting up the virtual entities are particularly evident when modern digital twin (DT) technology is to be used, as DT attempts to interactively mirror the \emph{real} world.

DT connects simulation with reality by requiring real-time bi-directional data exchange between virtual and physical entities.
The DT approach enables simultaneous development and testing on the virtual model and the real vehicle.

Increasing the number of parameters to describe reality in the simulation improves the simulation's credibility.
However, in the case of rapid development and testing, the important factor is the difficulty of creating virtual entities.
Therefore, the developers creating virtual entities manually to benefit from the DT approach must balance between a higher number of parameters and the development speed.
Automation helps to remove this trade-off and achieve both.

In this article, we argue for the automatic creation of virtual environments and vehicle models for ADS development.
We present use cases where it makes sense to prioritize simplicity of creation at the expense of the number of parameters to describe reality.
We present our DT generator\footnote{\url{https://git.sr.ht/~qeef/gen-sdf}}, which generates virtual entities, and use the generated virtual environment with the Gazebo simulator and the Robot Operating System (ROS)\footnote{\url{https://libre.video/videos/watch/e15b96af-9c2c-4b72-adf8-234148d6044c}}\footnote{\url{https://libre.video/videos/watch/b2e90dd6-ef61-4756-822f-a915aa2aa23a}}.
Particularly, the contribution of our work is:
\begin{itemize}
	\item We outline the relevance of DT to ADS development and testing.
	\item We present a low fidelity DT generator and showcase the generator output in the Gazebo Classic simulator.
	\item We discuss the use cases for a low fidelity DT.
\end{itemize}

In \cref{s:rw} we give an overview of the work related to ADS development, DT and simulators.
In \cref{s:gen} we present our generator and look at emerging possibilities brought by it.
In \cref{s:use} we discuss the situations in which ADS benefit from low fidelity DT.
We conclude our considerations in \cref{s:concl}.

\section{Related work}
\label{s:rw}

\begin{figure}[t]
	\centering
	\includegraphics[width=\linewidth,alt={Undirected graph showing the connection between the topics from the related work. The main connection is between the automated driving system and simulation nodes. To the simulation node, the following nodes are connected: fidelity, accuracy, co-simulation, and virtual entities. To the automated driving system node, the following nodes are connected: infrastructure, testing, development, vehicle, and reality. To the reality node, the physical entities node is connected. There are two more connections in the graph: the connection between the simulation and reality nodes is called reality gap; the connection between the virtual entities node and physical entities node is called twinning.}]{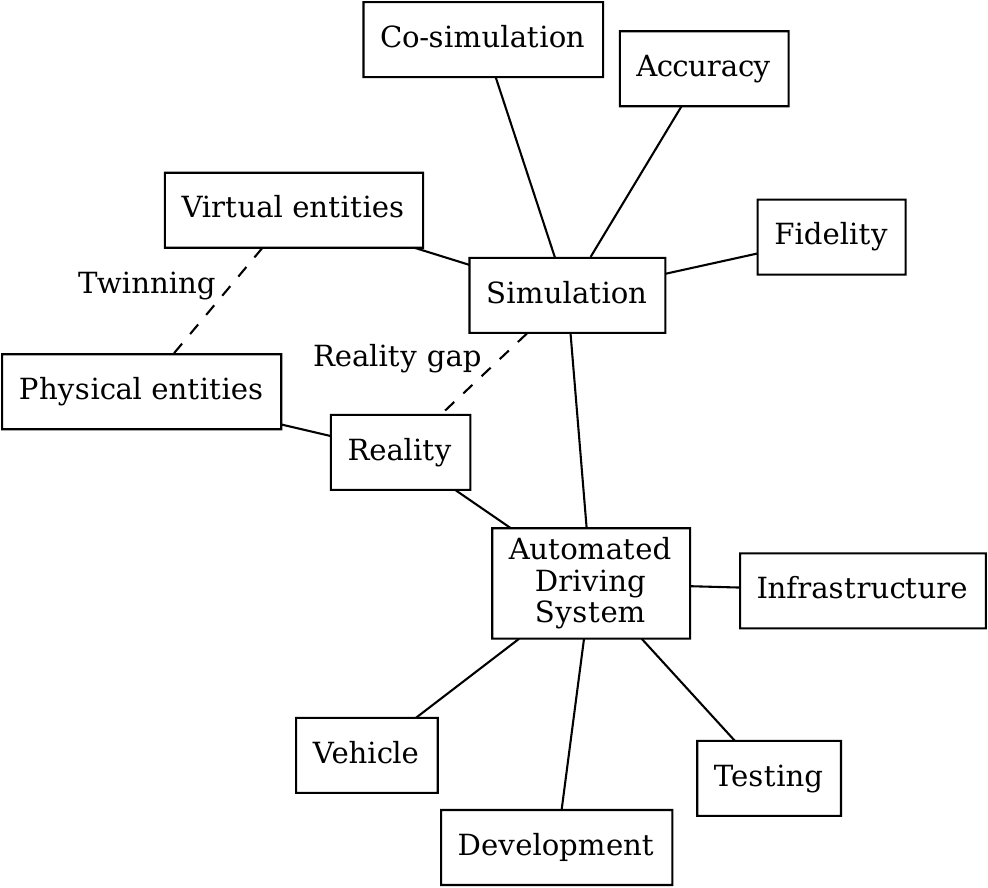}
	\caption{Connection between the topics of work on automated driving systems from the point of view of digital twin technology. The solid lines represent a close relevance. The dashed lines represent the connection between virtual and physical entities.}
	\label{f:rw-mm}
\end{figure}


Given the rapid progress in the development of automated driving systems (ADS), we can recognize similar efforts in the tools that promise to further accelerate the development of ADS.
\citet{macenski_robot_2022} introduce one such tool: the Robot Operating System (ROS) -- a broad ecosystem of libraries for robotics applications with a common consensus on interfaces.
ROS is the de facto standard for the development of robotics applications, including ADS; we focus on the tools that are compatible with ROS.

In \cref{s:rw:sim}, we provide an overview of simulators, a prominent group of tools for ADS development and testing.
In \cref{s:rw:dt}, we take a look at the use of digital twin (DT) technology for ADS.
In \cref{s:rw:dev}, we glance over the work in which DT is used for ADS development and testing, even if it is not explicitly mentioned.

\cref{f:rw-mm} depicts the connection between the topics of related work.
Lower part of the figure shows that real-vehicle development and testing is necessary for ADS.
Upper part of the figure shows that accuracy and fidelity are important factors of simulations.
Middle part of the figure represents how the simulation is connected with reality from DT point of view.

\subsection{Simulators}
\label{s:rw:sim}

\citet{kaur_survey_2021} review simulators for ADS testing and conclude that CARLA~\citep{Dosovitskiy17} and LGSVL~\citep{rong_lgsvl_2020} are the most suitable.
Gazebo~\citep{koenig_design_2004} is popular, but it is hard to create new environments for it.

CARLA was released as a completely new simulator developed from scratch for ADS research.
With a focus on the perception module, CARLA is an open source layer built on the highly realistic Unreal Engine\footnote{\url{https://www.unrealengine.com/en-US}}.
CARLA also features a ROS bridge to connect to ROS-based control algorithms for ADS.

LGSVL, on the other hand, was released with the main feature of smooth integration into the Autoware~\citep{kato_autoware_2018} -- open source ADS framework built on the ROS.
The development of LGSVL has been discontinued, but its successor, AWSIM\footnote{\label{fn:awsim}\url{https://tier4.github.io/AWSIM/}}, also provides first-class support for Autoware.

Gazebo focuses on robotics in general, not just ADS development.
The environment and robot models are specified in an SDFormat\footnote{\url{http://sdformat.org/}} file with XML syntax.
SDFormat is also supported by the new graphics engine o3de\footnote{\label{o3de}\url{https://o3de.org/}}.

\citet{koenig_design_2004} published Gazebo's original work in 2004 with case studies about reverse engineering real robots by creating the model in the simulator and simulating the robots in a virtual copy of the environment before deploying them. They even mentioned the possible real-time sensor data exchange between real and virtual entities. Then, almost ten years after the publication of Gazebo, \citet{shafto2012nasa} introduced DT.

\subsection{Digital twin technology}
\label{s:rw:dt}

In recent years, DT has also become increasingly important in the field of ADS.
\citet{schwarz_role_2022} describe what DT brings to ADS: ``The presence of communication [between virtual model and real vehicle] is a key differentiator for the digital twin.''
\citet{wright_how_2020} explain the difference between a digital twin and a model in simulation: ``A digital twin without a physical twin is a model.''
\citet{jones_characterising_2020} give a detailed characterization of the DT and name its most important components.

We emphasize two DT components important for our article: \emph{twinning} and \emph{fidelity}.
Twinning is a process of synchronizing parameter values between physical and virtual entities.
Fidelity describes the level of details in the simulation, including the number of twinned parameters.
Be aware of the difference between fidelity and \emph{accuracy}.
Fidelity tells us how detailed the virtual entity is compared to the physical one.
Accuracy describes the deviation of the twinned parameter values between the virtual and the physical entity.

The meanings of DT fidelity and fidelity of simulators overlap but are not the same.
In both cases, increasing fidelity improves credibility.
DT fidelity includes the number of \emph{twinned parameters}, and the fidelity of simulators depends on the number of \emph{simulated parameters}.

\citet{samak_autodrive_2023} present DT with all the necessary components. This includes physical and virtual entities for the environment (roads, traffic lights, smart infrastructure) and a real prototype vehicle (1:14). Their AutoDRIVE platform is based on ROS, has a Unity-based simulator and is intended for ADS research and education. The downside is the self-written simulator and the limited number of environments and vehicles.

\citet{stocco_mind_2023} describe how ADS testing in simulation relates to ADS testing in reality, targeting deep neural networks. They attempt to answer the question of whether the difference between virtual tests and real-world tests, the \emph{reality gap}, affects test performance. They conduct tests on a 1:16 prototyping platform that includes a high-fidelity simulator and conclude that real-world vehicle testing is still important. In other words, the realism of the simulators is not yet good enough.

\citet{hu_how_2023} provide an overview of work on the reality gap with the aim on perception and high-fidelity simulation. They divide the related topics into (i) knowledge transfer from simulation to reality, (ii) learning in digital twins, and (iii) learning through parallel intelligence. They conclude that there is always a reality gap and that there is a need to further investigate the methods for knowledge transfer.

From the work above, we could conclude that decreasing the reality gap requires increasing the fidelity of simulators.
This statement holds for the research in perception but not for ADS and DT in general.
We argue that it is important to consider DT fidelity, too.
Depending on the use case, we can achieve a sufficiently small reality gap even with a low-fidelity DT and simulator.

%


\subsection{Development and testing}
\label{s:rw:dev}

In this section, we review how various teams use DT for ADS development and testing.
\citet{wang_reduction_2021} evaluate ADS safety. First, they run ADS in parallel with the human driver but without access to the actuators and compare online the trajectories of the human driver with the trajectories generated by ADS. Later, they reduce uncertainties of the world model and carry out offline safety assessments of the recorded trajectories. Their goal is the offline safety assessment. The first stage, the online trajectory comparison, uses the elements of DT.

\citet{genevois_augmented_2022} inject lidar data at sensor level in real time from the simulation to the real ADS. This augmented reality approach has potential for vehicle-in-the-loop tests, especially when complex scenarios have to be tested on the real vehicle.

\citet{gelbal_vehicle_2022} describe a very similar approach, which they call \emph{vehicle in virtual environment}. They include ADS in the simulation with a prepared scenario and let the ADS react to this scenario. Both studies show the convergence of virtual and physical entities, which corresponds to DT.

Finally, \citet{drechsler_mire_2022} had several real road participants at different physical locations encounter each other in the simulation.
They demonstrate the concept of emergency braking system testing.
Looking from the DT perspective, ADS and the pedestrian synchronize their parameters, e.g., position and speed, to their virtual models.
ADS and the pedestrian also observe another participant via the simulation.
The result of the simulation is whether the collision happened.

This work show an attractive use of DT for ADS testing. It proves that what DT contributes to ADS is valuable and that there are use cases where high fidelity is not the most important thing. However, there is still room for automation in setting up DT, as DT is usually created manually, uses a virtual environment that only resembles a real environment, or uses virtual models that do not correspond to the real vehicles under test.

\section{Digital twin generator}
\label{s:gen}

In this section, we briefly introduce our digital twin (DT) generator for automated driving systems (ADS)\footnote{\url{https://git.sr.ht/~qeef/gen-sdf}}.
The generator produces the SDFormat\footnote{\url{http://sdformat.org/}} file that can be loaded into the Gazebo Classic, Gazebo, and o3de simulators, all providing first-class support for the Robot Operating System (ROS).
Thus, the generator's output enables ADS development and testing in DT fashion by setting up the simulator with (i) virtual entities, i.e., the environment and the models, and (ii) the interface to communicate with the entities via the ROS. We begin with reflections on the choices in developing the generator, showcase the generator output, and outline future work.

Our decisions are influenced by the fact that we strive to reflect the real world, consider DT with limited fidelity, and aim for simple DT creation.

\begin{figure}[t]
	\centering
	\includegraphics[width=\linewidth,alt={Directed graph showing the steps necessary to use data from the OpenStreetMap (first source) and vehicle parameters (second source) in the simulators CARLA (first sink), AWSIM (second sink), Gazebo Classic (third sink), and Gazebo (fourth sink). From OpenStreetMap, the arrow goes directly to CARLA, or to OpenDRIVE and then to CARLA, or to CommonRoad, Lanelet2, Unity editor, and then to AWSIM. From vehicle parameters, the arrow goes to Unreal Engine editor and then to CARLA, or to Unity editor and then to AWSIM. There are two more arrows from OpenStreetMap and vehicle parameters to SDFormat that represent the digital twin generator. The arrows continue from SDFormat to Gazebo Classic and Gazebo.}]{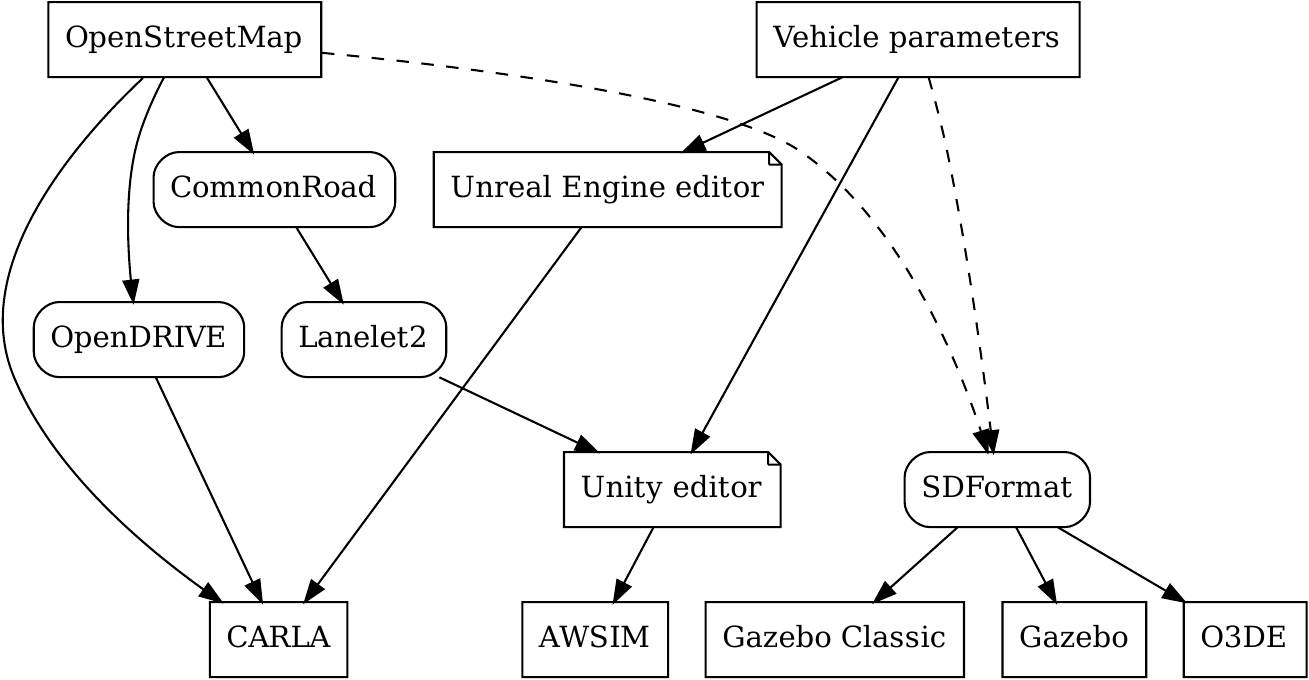}
	\caption{Diagram of the steps required to use OpenStreetMap data and vehicle parameters (boxes in the top row) in different simulators (boxes in the bottom row). The solid lines represent automatic conversions and the loading and saving of files in different formats (rounded boxes). The boxes for the Unreal Engine editor and the Unity editor represent non-trivial manual work. The dashed lines represent our generator.}
	\label{f:gen}
\end{figure}

To automatically create the virtual environment, we use OpenStreetMap\footnote{\url{https://www.openstreetmap.org/about}}, which provides detailed open map data for the whole world.
In \cref{f:gen}, we can see that there are tools for automatic conversion from OpenStreetMap data into OpenDRIVE~\citep{asam_ev_asam_2023}, CommonRoad~\citep{maierhofer_commonroad_2021} or Lanelet2~\citep{poggenhans2018lanelet2} formats.
However, the results of the conversions must be revised manually, as the conversion of detailed OpenStreetMap data to other formats involves many corner cases.
We limit corner cases by limiting the number of parameters we use to create the virtual environment.

To automatically create virtual models of the vehicles, we are left with the SDFormat and the Gazebo simulator.
In \cref{f:gen}, we can see that adding a new vehicle model to CARLA or AWSIM requires manual work in the Unreal Engine or Unity editors;
the dashed lines indicate the automatic conversions performed by our generator.

Our generator is a single-file Python 3 script that depends only on the standard library packages; this approach increases usability in different environments.
We set the virtual environment's bounding box in the script file.
The environment consists of buildings (coordinates and height estimation) and roads (coordinates of centerline between two lanes with fixed width).

We also set the vehicle names and types in the script file, where the type is \emph{twin} for the vehicle model based on \citep{GazeboFuel-OpenRobotics-Prius-Hybrid} with  GPS sensor and Ackermann steering capable of bi-directional data exchange with the real vehicle, \emph{shadow} for the model that only listens to the GPS position from the real vehicle, and \emph{ghost} for a shadow that does not collide with the other models.

From our experience with CARLA, adding OpenStreetMap data is trivial manual work and about half an hour of processing with a dubious result. Adding a custom vehicle model is non-trivial manual work and hours of compiling. Compared to our generator, the generator always uses OpenStreetMap data and the bounding box can be changed in the script file. Adding a custom vehicle model requires familiarity with SDFormat and more elaborate changes in the script file. With the default bounding box, the script finishes under a second and the result can be loaded into the simulator.

We tested the generated SDFormat file in the Gazebo Classic simulator by replaying the data recorded in ROS bag files: replaying GPS positions\footnote{\url{https://libre.video/videos/watch/e15b96af-9c2c-4b72-adf8-234148d6044c}} and also vehicle control messages\footnote{\url{https://libre.video/videos/watch/b2e90dd6-ef61-4756-822f-a915aa2aa23a}}.



Future work is to use our generator in the situations described in \cref{s:use}.
We also want to increase the fidelity of our generator; we plan to support more features of OpenStreetMap, such as traffic lights, to introduce bi-directional data exchange to the environment.
Finally, we plan to predefine more vehicle sensors and manage vehicle parameters more comfortably.

\section{Digital twin use cases}
\label{s:use}

\begin{figure}[t]
	\centering
	\includegraphics[width=\linewidth,alt={Four screenshots of Gazebo Classic simulator depicting a part of the test track in Weissach, Germany. The part of the track looks like an incomplete circle. The four screenshots differs in the positions of the color boxes shown on the track. The color boxes represent the movement of the vehicle model (green) and the vehicle shadow (blue). It looks that the green box is slightly delayed from the blue box.}]{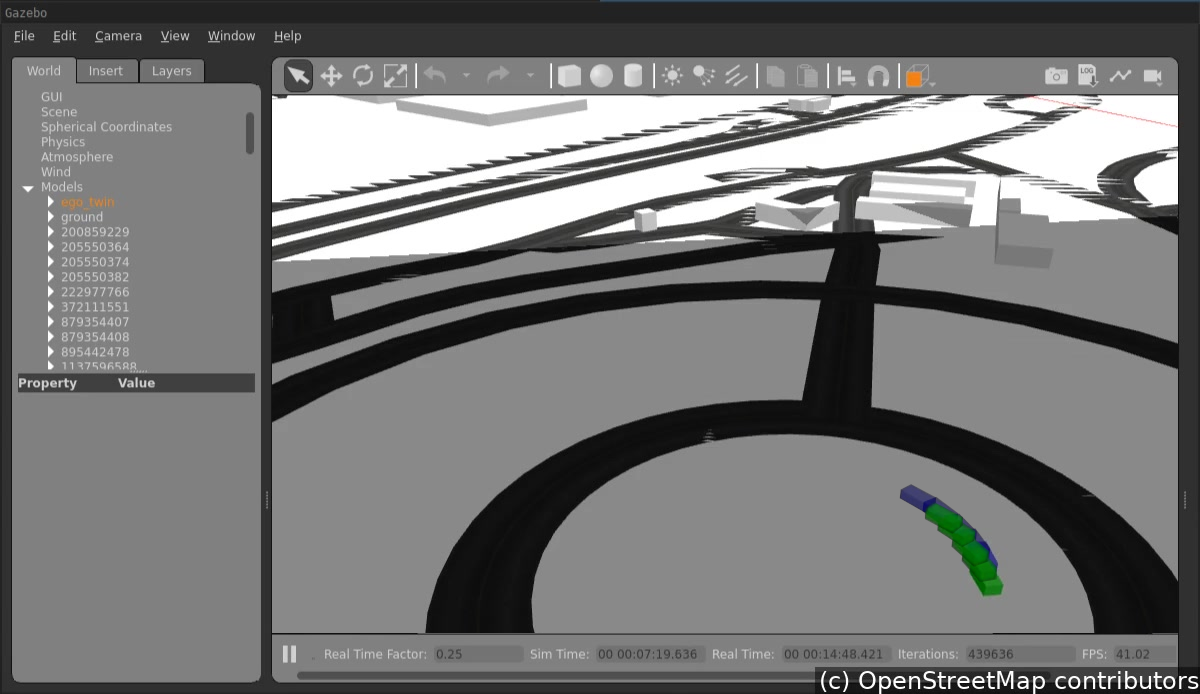}
	\caption{The screenshot of the Gazebo Classic simulator with automatically generated virtual entities reflecting the test track in Weissach, Germany, and the ADS-equipped vehicle under test. Buildings and roads {\small\copyright}~OpenStreetMap contributors.}
	\label{f:weissach}
\end{figure}

In this section, we present situations related to the development and testing of automated driving systems (ADS), where limited fidelity is not a major concern, but the automatic generation of components related to digital twin (DT) technology, i.e., virtual environments and vehicle models, is highly appreciated.
\cref{f:weissach} shows an example of generated virtual entities that communicate via Robot Operating System (ROS) messages.

When we start working on a real vehicle, we must first determine its parameters.
If the parameters are already available, they may be incorrect, so we have to adjust them.
The identification and tuning of the model is covered in \cref{s:use:id}.
When the vehicle model is known, we can observe the reality gap.
In \cref{s:use:val}, we discuss the deviation between the parameter values observed in the reality and simulation.
When the reality gap is known, we want to evaluate various ADS algorithms for trajectory planning, decision-making, and vehicle control for research problems in the field of automated driving, such as parking in the city, highway overtaking, racing, or crossing intersections, as outlined in \cref{s:use:alg}.

During the process, more parameters may need to be added, some adjusted, and others invalidated, so the cycle of identification, tuning, validation, and evaluation starts over again.
However, from the perspective of DT, the above steps are the same.
Using DT in development and testing means that we can constantly monitor the reality gap and immediately fix bugs that are discovered at any stage of the development and testing cycle.

\subsection{Online identification and tuning}
\label{s:use:id}

The goal of vehicle model identification and tuning is to determine or adjust the parameters of the vehicle model based on the ground truth data from the real vehicle.
This is important when ADS development starts with a new vehicle or when something happens to the vehicle that may affect the parameters, such as a tire change.
An example of parameters that need to be determined and adjusted is the mapping between the steering wheel angle and the corresponding front wheel angle at a certain speed.

Using DT helps to determine the parameters by requiring bi-directional real-time data exchange that presumes a compatible interface between the real vehicle and vehicle model.
The real-time nature of the communication enables instant estimation and adjustment of the vehicle model parameter values, either by hand or using optimization techniques.


\subsection{Observing the reality gap}
\label{s:use:val}

There is no ground truth when observing the reality gap; we want to indicate whether the mismatch is caused by the real vehicle or the vehicle model.

If the control works for the virtual model but fails for the real vehicle, we call the situation \emph{real vehicle bias}.
An example is the algorithm producing the steering angle request in degrees instead of radians.
This type of error is caused by lacking knowledge of the vehicle interface and is often observed in the initial phase of the experimentation.

If the control works for the real vehicle but fails for the vehicle model, we call the situation \emph{vehicle model bias}.
An example is a set of incorrect parameters of the vehicle model such as a smaller maximum steering radius than that of the real vehicle.
This type of error could be caused by incorrect parameter tuning of the vehicle model.

We must understand the reality gap before validating and evaluating ADS algorithms; when we know its limits, we can make conclusions about the ADS algorithm under the test.
DT is a great tool for this because the reality gap can be observed in real time.


\subsection{Driving through an intersection}
\label{s:use:alg}

Finally, we want to know whether an ADS algorithm developed in the simulation works in the real vehicle and how it performs.
The reality gap is known from previous experiments, and so is the deviation of the parameter values.

As a simple example, we study the validation and evaluation of a trajectory planning algorithm for roundabout intersection traversal.
While approaching the intersection, we can generate a virtual replica of the roundabout from real-world OpenStreetMap data and the vehicle model representing the ADS-equipped vehicle under the test.

Then, we can use a mixed-reality approach and simulate other road participants to test different scenarios in the real vehicle.
In parallel, we can run multiple simulations testing similar scenarios based on reality (sensor messages from the real vehicle) but representing different situations.
We observe the differences between trajectories planned in the simulation with the trajectories planned in reality in real time.
We observe the vehicle behavior in the simulation and reality in real time.

Using the DT generator, we can easily test the algorithms on multiple intersections during a single real-vehicle test drive.
Moreover, we can simulate multiple scenarios at each of the intersections at the same time.

\section{Conclusion}
\label{s:concl}

We look at automated driving systems, digital twin technology, and simulators from the perspective of development and testing with Robot Operating System.
We argue in favor of digital twins with limited fidelity and their appropriate use in the development and testing of automated driving systems.
We advocate automation in the creation of digital twins and show the benefits of this approach on an ADS for roundabout intersection traversal.

We present a generator for low fidelity digital twins and situations where it is advantageous to favor automated generation over high fidelity simulation.
We successfully tested the output of the generator in the Gazebo Classic simulator.

Future work with digital twin technology will focus on the investigation of non-reproducible or poorly reproducible events such as traffic accidents or crashes.
Realistic testing of these events is not possible on a large scale due to safety, financial, and other constraints; nevertheless, these events will certainly occur.
Automated driving systems should be verified and validated to deal with these situations properly.
Digital twin technology with the mixed reality approach is perfect for this.

Another future research direction is reinforcement learning for control systems, in which data from the real vehicle and several vehicle models are combined to increase learning performance.

\section*{Acknowledgement}
\label{s:ack}

This work was developed during an internship at Virtual Vehicle Research GmbH.
This work was supported by the Grant Agency of the Czech Technical University in Prague, grant No. SGS22/167/OHK3/3T/13.
This work was supported by the Technology Agency of the Czech Republic under the project Certicar CK03000033.
This work was co-funded by the European Union under the project ROBOPROX (reg. no. CZ.02.01.01/00/22\_008/0004590).

\bibliographystyle{plainnat}
\bibliography{main}
\end{document}